\pdfoutput=1

\documentclass[11pt]{article}

\usepackage[final]{acl}
\usepackage{algorithm}
\usepackage{algpseudocode}
\usepackage{CJK}
\usepackage{multirow}
\usepackage[OT2,T1]{fontenc}
\usepackage{colortbl}
\newcommand\textcyr[1]{{\fontencoding{OT2}\selectfont #1}}
\definecolor{LightGray}{gray}{0.9} 
\usepackage{arydshln}

\usepackage{times}
\usepackage{latexsym}
\usepackage{amsmath}
\usepackage[T1]{fontenc}

\usepackage[utf8]{inputenc}
\usepackage{amssymb}
\usepackage{microtype}

\usepackage{inconsolata}

\usepackage{graphicx}

\usepackage[inkscapelatex=false]{svg}

\title{\textit{Bias Beyond English}: Evaluating Social Bias and Debiasing Methods in a Low-Resource Setting}

\author{
 \textbf{Ej Zhou\textsuperscript{1,2}},
 \textbf{Weiming Lu\textsuperscript{2}}
\\
 \textsuperscript{1}Language Technology Lab, University of Cambridge\\
 \textsuperscript{2}College of Computer Science and Technology, Zhejiang University
\\
 \small{
   \texttt{yz926@cam.ac.uk, luwm@zju.edu.cn}
 }
}

\begin{document}
\maketitle

\begin{abstract}
\textit{This paper addresses the identification of social harms—including stereotypes and bias—which may be upsetting to some readers.}

Social bias in language models can potentially exacerbate social inequalities. Despite it having garnered wide attention, most research focuses on English data. In a low-resource scenario, the models often perform worse due to insufficient training data.  This study aims to leverage  high-resource language corpora to evaluate bias and experiment with debiasing methods in low-resource languages. We evaluated the performance of recent multilingual models in five languages: English (\textsc{eng}), Chinese (\textsc{zho}), Russian (\textsc{rus}), Indonesian (\textsc{ind}) and Thai (\textsc{tha}), and analyzed four bias dimensions: \textit{gender}, \textit{religion}, \textit{nationality}, and \textit{race-color}. By constructing multilingual bias evaluation datasets, this study allows fair comparisons between models across languages. We have further investigated three debiasing methods-\texttt{CDA}, \texttt{Dropout}, \texttt{SenDeb}-and demonstrated that debiasing methods from high-resource languages can be effectively transferred to low-resource ones, providing actionable insights for fairness research in multilingual NLP.
\end{abstract}

\section{Introduction}

Machine learning frameworks are fundamentally designed as a function that generalizes past data \citep{Chun}. As a result, pretrained language models inevitably learn the inherent social biases embedded in raw real-world text. Studies have shown that these models and their learned representations \textbf{retain} and \textbf{propagate} biases from their training data. For example, \citet{bolukbasi2016man} found that word embeddings such as \texttt{Word2Vec} retain measurable gender bias, with \textit{male}-associated terms
 being linked to professions like \textit{programmer} and \textit{scientist}, while  \textit{female}-associated terms are more commonly linked to \textit{nurse} and \textit{homemaker}. 
Meanwhile, bias is propagated in downstream applications, thus potentially reinforcing stereotypes and exacerbating information inequality \citep{bender2018data}.

Social bias has been extensively studied in natural language processing \citep{vanmassenhove2019lost, kiritchenko2018examining, sap2019risk,nangia2020crows,nadeem2020stereoset}. However, most of the relevant works are limited to English and reflect an Anglo-centric social context.  The ``\textit{\textit{subaltern}}'', a key figure in postcolonial discourse, is historically marginalized and voiceless—``\textit{has no history and thus no voice}'' \cite{morris2009can}.
In computational linguistics, this \textbf{voiceless}ness of the subaltern is also pronounced: Most NLP research overlooks thousands of languages spoken by billions of people \citep{bender2018data, eberhard}.
 Although multilingual language models are trained on language data rather than cultural data, all languages inherently reflect cultural stereotypes. 
 However, models are primarily trained, evaluated, and aligned using Western data and culture, and therefore debiasing techniques often fail to account for culture-specific discrimination \citep{khandelwal2023casteist}. The systemic bias present in these models not only affects the fairness and accuracy of multilingual systems, but also negatively impacts social equity and equal access to information. Developing more equitable and inclusive multilingual systems is an urgent challenge in NLP today.

This study aims to systematically evaluate model bias in a multilingual setting. We adopt the masked language model prediction probability method proposed by \citet{nadeem2020stereoset} to measure the bias of language models toward specific social attributes. To compare biases across different models and languages, we propose a new evaluation metric that standardizes model evaluation indicators, enabling more precise cross-model bias comparison.

Based on the CrowS-Pairs dataset \cite{nangia2020crows}, we created a new multilingual bias evaluation dataset with English, Chinese, Russian, Thai, and Indonesian; each with four representative bias types: gender, race-color, nationality and religion. These languages were selected based on their global use and the distribution of online resources, with Thai and Indonesian considered low-resource languages. In our experiments, we evaluated a series of widely used multilingual language models. We noticed significant differences in the type of model bias across languages. Our findings emphasize the importance of incorporating diverse cultural and linguistic backgrounds into research to ensure fairness on a global scale.

Little research has been done on exploring application in debiasing methods across languages. In this work, we further fine-tuned multilingual models on English Wikipedia datasets with \texttt{CDA}, \texttt{DO}, computed the bias subspace with \texttt{SenDeb}, and measured bias shifts across English and four other languages. 

We summarize our contribution as follows.
\begin{itemize}

\item we proposed $\mathbb{NBS}$ (\S \ref{sec: NBS} \& \S \ref{sec:NBS_definition}) as a method to evaluate model bias in a multilingual setting by measuring the normalized probability of a masked-word prediction in a biased context.
\item  we curated a new dataset (\S \ref{sec: dataset_construction}) for bias evaluation with five languages, representatively selected based on the language resource conditions, with Thai and Indonesian considered low-resource languages.
\item our results show that the impact of different bias types (\emph{e.g.}, gender, religion) varies across languages and cultures in the six multilingual models we tested (\S \ref{sec: results}).
\item we demonstrated that debiasing strategies can be effectively transferred to other languages through cross-lingual knowledge sharing (\S \ref{sec: debias_reuslt}).

\end{itemize}

\section{Related Works}

\subsection{Low-Resource Languages}
Low-resource language communities often face barriers accessing information. More research on low-resource languages will help develop NLP tools that allow these communities to access and share information equitably, reducing the digital divide and promoting social inclusion, and will also contribute to developing more universal cross-linguistic models. \citet{hedderich-etal-2021-survey} discussed how low-resource languages can benefit from annotated resources in high-resource languages. 

However,  \citet{pmlr-v119-hu20b} and \citet{wu-dredze-2020-languages} noted that there remains a significant performance gap between high-resource and low-resource settings. Despite advancements in multilingual NLP, existing models do not yet serve as truly universal language models, and many languages with over a million speakers remain underrepresented \citep{lauscher-etal-2020-araweat}.

\subsection{Social Bias in the Multilingual Setting}

A significant body of research has emerged on bias in NLP systems. The attempts to measure bias in word embeddings date back \citet{bolukbasi2016man}, followed by research exposing bias in contextualized language representations trained for various NLP tasks. For example, \citet{vanmassenhove2019lost} identified bias in machine translation, \citet{kiritchenko2018examining} found gender and racial bias in sentiment analysis , and \citet{sap2019risk} discovered racial bias in hate speech and toxicity detection. However, these are all limited to the English language.

Some research explored social biases in non-English settings. \citet{lauscher-etal-2020-araweat} conducted an extensive analysis of bias in Arabic word embeddings, identifying gender and racial biases in Arabic news corpora. \citet{sahoo-etal-2023-prejudice} created a Hindi social bias detection dataset. \citet{neveol-etal-2022-french} extended CrowS-Pairs to investigate various biases in French. \citet{zhou-etal-2019-examining} evaluated gender bias in grammatically gendered languages, with experiments on French and Spanish text. \citet{b-etal-2022-casteism} examined gender and caste bias in monolingual word embeddings for Hindi and Tamil. Their research demonstrated that bias evaluation becomes significantly more complex in a multilingual context due to (1) varying cultural frameworks that influence the definition of bias and (2) differences in grammatical structures, which render some existing evaluation methods highly challenging to apply.

Recently, several studies have begun to examine multilingual bias at a more holistic level.
\citet{ahn-oh-2021-mitigating} analyzed ethnicity bias in six languages and attempted to mitigate biases seen in monolingual models by using mBERT.
\citet{levy-etal-2023-comparing} analyzed biases in sentiment analysis in five languages in mBERT and XLM-RoBERTa. \citet{camara-etal-2022-mapping} analyzed gender, race, and ethnicity bias in English, Spanish, and Arabic for the sentiment analysis task. 
\citet{cabello-piqueras-sogaard-2022-pretrained} created parallel cloze test sets in English, Spanish, German and French with mBERT, XLM-R and m-T5. However, all the aforementioned studies have exclusively utilized comparatively small-scale pre-trained language models and have not examined bias behavior in large language models, nor have they focused on low-resource languages.

\begin{table*}
    \centering
    \small
    \begin{tabular}{lcccccc}
        \hline
        \textbf{Language} & \textbf{Language Family} & \textbf{Writing System} & \textbf{Availability} & \textbf{\texttt{2023-50}} & \textbf{\texttt{2024-10}} & \textbf{\texttt{2024-18}} 
        \\ \hline
        English  & Indo-European (Germanic) & Latin & High & 44.43\% & 46.45\% & 45.51\% \\ 
        Chinese & Sino-Tibetan & Hanzi & Medium & 5.08\% & 4.17\% & 4.42\% \\ 
        Russian  & Indo-European (Slavic) & Cyrillic & Medium & 6.03\% & 5.81\% & 5.95\% \\ 
        Thai & Kra-Dai & Thai & Low & 0.43\% & 0.41\% & 0.41\% \\ 
        Indonesian  & Austronesian & Latin & Low & 0.86\% & 0.86\% & 0.92\% \\ \hline
    \end{tabular}
    \caption{Linguistic characteristics of selected languages from the Common Crawl archives \cite{CommonCrawlLanguages}. Each entry corresponds to the dataset prefix \texttt{CC-MAIN}.}
    \label{tab:language-characteristics}
\end{table*}

\subsection{Debiasing in Multilingual Systems}

Existing debias methods have made progress in reducing biases in models, but still face numerous challenges. \citet{meade2022empirical} summarized several recent debias methods for pretrained language models including
Counterfactual Data Augmentation (\texttt{\textbf{CDA}}) \citep{zmigrod2019counterfactual, webster2020measuring}, \textbf{\texttt{Self-Debias}} \citep{schick2021selfdiagnosis}, Dropout Regularization (\textbf{\texttt{DO}}), Sentence-Level Debiasing (\textbf{\texttt{SenDeb}}) \citep{liang2020towards} and Iterative Nullspace Projection (\textbf{\texttt{INLP}}) \citep{ravfogel-etal-2020-null}.

Most debias methods are designed and evaluated for English-only environments, and research on multilingual transfer learning for debiasing is still limited. Since multilingual models share linguistic knowledge across languages, they have the potential to transfer debiased knowledge from English to other languages \citep{wang-etal-2019-compact}. \citet{reusens2023investigatingbiasmultilinguallanguage}
explored the cross-lingual transferability of debiasing techniques in multilingual models using mBERT, demonstrating that debiasing methods effectively reduce bias with SentenceDebias achieving the best results. \citet{nozza-2021-exposing} explored cross-lingual debiasing in English, Italian, and Spanish for stereotype detection tasks. However, no current research focuses on debiasing specifically for low-resource languages, highlighting an urgent need for multilingual debiasing transfer learning.

\section{Multilingual Bias Evaluation}

\subsection{Methodology}
\label{sec: NBS}

Adopting the methodology proposed by \citet{nangia2020crows}, we also assess bias in masked language models (MLM) by predicting the probability of masked words with pseudo-log-likelihood estimation \citep{wang2019bert}. The probability score $\mathbb{PS}({s_i})$ of sentence $s$ is defined as:

\begin{equation}
   \mathbb{PS}({s_i}) = \sum_{j=0}^{|U|} \log P(u_j \in U | U_{\backslash u_j}; m_i; \theta) 
\end{equation}

where $M = \{m_{0},\ldots,m_{n}\}$ represent the stereotypical words we modify in this case, $U = \{ u_0,\ldots,u_l\}$ represent the unchanged words, $s_i = U \cup m_i$, and $\theta$ is the language model parameter. Most recent multilingual models are causal language models (CLM) which are not finetuned to predict masked tokens, therefore, instead of masking a token, we remove it from the input and use the model to generate a probability distribution for that position. We take the prediction scores of the language modeling head (scores for each vocabulary token before SoftMax) as an approximation.

While \citet{nangia2020crows}’s method allows for intra-model comparisons, it does not facilitate bias comparisons across different models. Therefore, we define Normalized Bias Score $\mathbb{NBS}$ (\ref{equ:NBS_main}) for comparison across models and provide a benchmark framework:

\begin{equation}
\label{equ:13}
W_{\text{avg}} = \frac{1}{n} \sum_{l \in \text{lang}} \frac{1}{N} \sum_{k=1}^{N} \frac{\left| \mathbb{PS}(s_{l,k}) + \mathbb{PS}(\bar{s}_{l,k}) \right|}{2}
\end{equation}

\begin{equation}
\label{equ:NBS_main}
\begin{split}
\mathbb{NBS}(\theta) & = \frac{1}{W_{avg}}\cdot \frac{1}{N} \sum_{k=1}^{N} \left| \mathbb{PS}({s_{l,k}}) - \mathbb{PS}({\bar{s}_{l,k}}) \right| \\ \\
           & = 2n \cdot 
           \frac
           {\sum_{k=1}^{N} \left| \mathbb{PS}({s_{l,k}}) - \mathbb{PS}({\bar{s}_{l,k}}) \right|}
           {\sum_{l\in lang} \sum_{k=1}^{N} \left| \mathbb{PS}({s_{l,k}}) + \mathbb{PS}({\bar{s}_{l,k}}) \right|}
\end{split}
\end{equation}

where $n = \left| \text{lang} \right| $, $s$ the original sentence and $\bar{s}$ the modified. In our analysis, bias evaluations are conducted using $\mathbb{NBS}$ metric. The closer $\mathbb{NBS}$ is to 0, the lower the bias in the model. If $\mathbb{NBS}$ = 0, it indicates that the model treats the two terms equally and exhibits no intrinsic social bias. For a complete mathematical illustration of this section, please refer Appendix \ref{sec:NBS_definition}.

\subsection{Dataset Construction}

\label{sec: dataset_construction}

To evaluate bias in low-resource languages, we build our dataset based on CrowS-Pairs \citep{nangia2020crows}. 

The dataset contains 1,508 examples,
covering various bias types 
 and measures model bias by comparing the likelihood of stereotypical \emph{vs.} non-stereotypical sentences. 
Previous research shows that many widely used language models favor stereotypical sentences in English, revealing internal biases.
However, CrowS-Pairs is based on American sociocultural contexts, which limits its applicability to other languages and cultures. To make bias evaluation more representative, this study selects four primary bias types for in-depth analysis: \textit{Gender, Race-Color, Nationality, Religion}, which are relatively generalizable across different cultural and linguistic contexts.
After filtering the dataset, we obtained 1,042 sentences, adapted for evaluating bias in non-English languages.

We translated CrowS-Pairs into four languages: Chinese (\textsc{zho}), Russian (\textsc{rus}), Indonesian (\textsc{ind}), and Thai (\textsc{tha}) using the Google Translate API. The languages were chosen based on Common Crawl \cite{CommonCrawlLanguages}, a large-scale web dataset that reflects the availability of online resources in different languages. Table \ref{tab:language-characteristics} shows the proportion of web content in these languages. According to linguistic resource classifications in NLP \citep{joshi-etal-2020-state}, Chinese and Russian are considered high-resource, while Thai and Indonesian are low-resource languages, despite having millions of native speakers. The translation tool was chosen considerting both the effectiveness and budget-resource constraints. Despite the resulting corpus being artificial and translation-based, we will see in the later section that the experimental results still reveal meaningful patterns, demonstrating the value of the evaluation albeit limitations.

\begin{table*}
    \small
    \centering
     \resizebox{.94\textwidth}{!}
     {
    \begin{tabular}{l l cccccc}
        \hline
        \multirow{2}{*}{Language} & \multirow{2}{*}{Bias Type} & \multicolumn{6}{c}{Bias Score per Model} \\ 
        & & \textbf{mBERT} & \textbf{XLM-R} & \textbf{XGLM} & \textbf{Gemma 3} & \textbf{Qwen 2.5} & \textbf{LLaMA 3} \\ \hline \hline 
        \multirow{5}{*}{English} & Gender & 52.57 & 41.31 & 20.42 & 18.98 & 11.99 & 7.53 \\
          & Nationality & 43.37 & 36.87 & 16.24 & 14.23 & 9.90 & 7.45 \\
          & Race-color & 44.51 & 33.54 & 16.24 & 17.86 & 11.67 & 7.50 \\
          & Religion & 48.49 & 44.07 & 18.97 & 16.80 & 11.16 & 9.28 \\
          \noalign{\vskip 1pt} 
          \cdashline{2-8}
          \noalign{\vskip 2pt} 
          & \textbf{Average} & \textbf{46.76} & \textbf{37.06} & \textbf{17.57} & \textbf{17.48} & \textbf{11.43} & \textbf{7.68} \\
        \hline
        \multirow{5}{*}{Chinese} & Gender & 58.19 & 43.24 & 25.00 & 26.67 & 25.76 & 11.46 \\
          & Nationality & 59.35 & 47.46 & 20.49 & 28.96 & 29.24 & 10.94 \\
          & Race-color & 56.79 & 55.71 & 21.01 & 25.84 & 25.25 & 11.04 \\
          & Religion & 60.49 & 50.54 & 18.81 & 25.71 & 29.35 & 10.16 \\
                    \noalign{\vskip 1pt} 
          \cdashline{2-8}
          \noalign{\vskip 2pt} 
          & \textbf{Average} & \textbf{57.91} & \textbf{50.79} & \textbf{21.71} & \textbf{26.51} & \textbf{26.40} & \textbf{11.04} \\
        \hline
        \multirow{5}{*}{Russian} & Gender & 53.27 & 37.11 & 26.42 & 40.91 & 14.77 & 11.24 \\
          & Nationality & 49.59 & 37.13 & 19.44 & 25.86 & 10.89 & 9.45 \\
          & Race-color & 48.79 & 39.42 & 25.48 & 29.49 & 13.07 & 10.51 \\
          & Religion & 46.65 & 37.61 & 31.16 & 25.09 & 12.53 & 10.76 \\
                    \noalign{\vskip 1pt} 
          \cdashline{2-8}
          \noalign{\vskip 2pt} 
          & \textbf{Average} & \textbf{49.82} & \textbf{38.30} & \textbf{25.37} & \textbf{31.36} & \textbf{13.11} & \textbf{10.56} \\
        \hline
        \multirow{5}{*}{Indonesian} & Gender & 59.42 & 38.72 & 17.58 & 14.91 & 24.95 & 8.62 \\
          & Nationality & 62.83 & 50.43 & 12.94 & 12.73 & 21.23 & 7.61 \\
          & Race-color & 57.07 & 55.38 & 19.64 & 14.90 & 23.69 & 10.02 \\
          & Religion & 60.84 & 55.08 & 24.93 & 20.68 & 33.23 & 8.37 \\
                    \noalign{\vskip 1pt} 
          \cdashline{2-8}
          \noalign{\vskip 2pt} 
          & \textbf{Average} & \textbf{58.92} & \textbf{50.40} & \textbf{18.63} & \textbf{15.16} & \textbf{24.59} & \textbf{9.13} \\
        \hline
        \multirow{5}{*}{Thai} & Gender & 74.84 & 55.15 & 20.92 & 33.34 & \cellcolor{LightGray}13.01 & 11.51 \\
          & Nationality & 103.26 & 65.37 & 22.04 & 25.94 & \cellcolor{LightGray}11.67 & 7.52 \\
          & Race-color & 87.08 & 64.93 & 21.41 & 30.68 &\cellcolor{LightGray}12.75 & 10.21 \\
          & Religion & 128.99 & 73.60 & 25.57 & 40.94 &\cellcolor{LightGray}15.88 & 11.02 \\
                    \noalign{\vskip 1pt} 
          \cdashline{2-8}
          \noalign{\vskip 2pt} 
          & \textbf{Average} & \textbf{90.69} & \textbf{63.41} & \textbf{21.80} & \textbf{31.66} & \cellcolor{LightGray}\textbf{12.96} & \textbf{10.21} \\
        \hline
  \hline  
  
  \end{tabular}
  }

    \caption{Bias Score Comparison of Different Models Across Languages}

    \label{tab:bias-scores}

\end{table*}

\subsection{Experiment}

\subsubsection{Setting}

In this study we selected six pre-trained models: mBERT, XLM-RoBERTa, XGLM, Gemma 3, Qwen 2.5 and LLaMA 3. These models are widely used particularly in multilingual tasks. For the details of the models, refer Appendix \ref{sec:models}. Note that all language models in this list support all evaluated languages, except Qwen 2.5, which has reported to support the other four language but not Thai, as indicated in its technical documentation \citep{qwen2025qwen25technicalreport}. We still report the bias scores calculated for Qwen 2.5 in Thai as per our evaluation methodology; however, the reader should be advised to keep the information in mind that Qwen 2.5 model lacks proper support for Thai language, as will be discussed in later sections.

All experiments were conducted on NVIDIA A100 PCIE GPUs.
To ensure fair comparison, all models were tested under identical hardware and software conditions.

\subsubsection{Results \& Analysis}

\label{sec: results}

\begin{figure*}[t]
  \centering
  \includegraphics[width=1.03\linewidth]{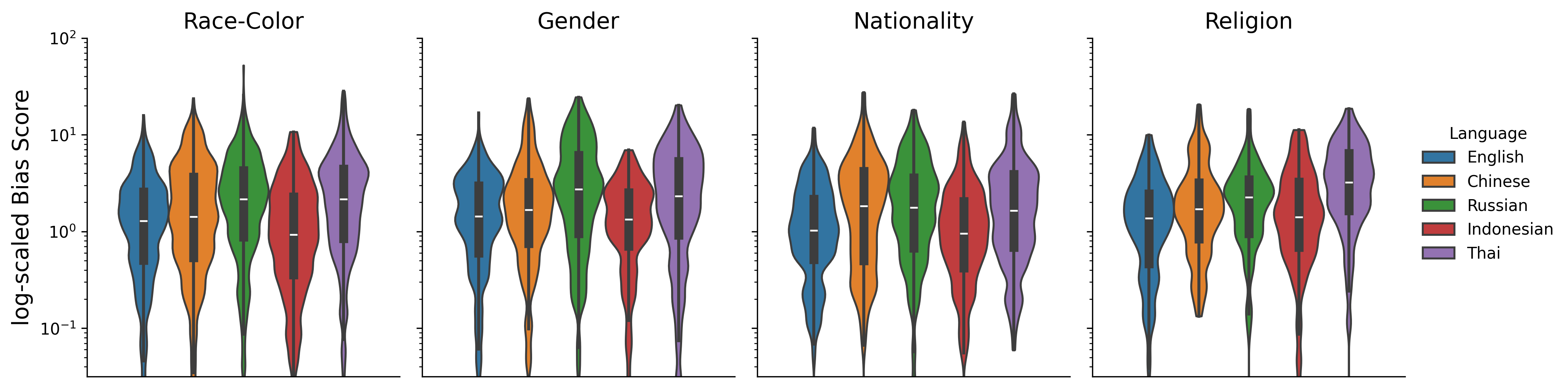}
    \caption {Bias Score across different languages with different bias categories in Gemma.}
  \label{fig:bias_violin_mbert}
  
\end{figure*}

We first ran inference on 1,042 dataset samples to obtain key bias evaluation metrics:
$\mathbb{PS}({s})$ and $\mathbb{PS}(\bar{s})$, which represent the model's bias toward specific social attributes.
We then applied Equation \ref{equ:NBS_main} to compute bias scores across different scenarios.
The results are detailed in Table \ref{tab:bias-scores}, which provides an overview of the bias scores for different models and languages.  Figure \ref{fig:bias_violin_mbert} visualizes the results of the Gemma model, illustrating the bias distribution across different languages and social attributes. To view all graphical results, please refer to Appendix \ref{sec:appendix-graph}.

Through our classification of bias types, we observe notable differences across different experimental settings. Some potential insights are as follows:

\paragraph{Model-wise}
Smaller models like XLM-RoBERTa and mBERT exhibit the highest bias scores among the tested models across multiple languages, particularly in Thai (90.69). XGLM, Gemma and Qwen 2.5 demonstrate moderate bias scores, often lower than XLM-RoBERTa and mBERT but still displaying noticeable biases across languages. LLaMA 3 consistently displays the lowest bias scores across all categories in English, suggesting it incorporates stronger alignment strategies to mitigate social biases. The lower bias scores observed in LLaMA 3 align with recent advancements in alignment-focused training methodologies. Meta’s development of LLaMA models emphasizes reinforcement learning from human feedback (RLHF) and instruction tuning to ensure outputs adhere to ethical considerations and fairness principles \citep{grattafiori2024llama3herdmodels}. 

Interestingly, Qwen 2.5, despite being a highly performing model specifically in Chinese NLP benchmarks \citep{qwen2025qwen25technicalreport}, exhibits a strong bias in Chinese (26.40), whereas Qwen 2.5 produced a very low bias score in Thai. However, this does not suggest that the model has relatively less bias in Thai than in English, because the model’s lack of comprehension in Thai leads to unintelligible or generic outputs which resulted in a lower calculated bias score. In contrast its proficiency in Chinese allows for a finer comprehension of the contexts, which in turn exposes more detectable biases. This finding reinforces the idea that bias is independent of overall performance and can persist, if not more represented, in larger language models with stronger performances, without alignment interventions. This supports the idea that bias is not merely a function of model size or performance, but rather a reflection of data composition, pretraining strategies, and alignment interventions \citep{bender2021dangers}.

\paragraph{Language-wise} English has the lowest bias scores across all models (7.68 in  3); Russian, despite being a relatively high-resource language, has considerably high bias scores in XGLM and XLM-RoBERTa, particularly in gender and race-color. 
Chinese exhibits higher bias scores, despite being a highly resourced language. This may suggest that its character-based writing system and different sociocultural contexts contribute to increased biases. 
Indonesian scores better than expected, notably in XGLM (18.63) and  3 (9.13), possibly due to its simpler grammar and usage of Latin script, Thai exhibits the highest bias scores across models. This suggests that Thai, a low-resource language with a distinct script and structure, faces significant underrepresentation in training data, leading to increased biases.

\paragraph{On Religious Bias} 
Unlike the other languages, Thai and Indonesian consistently exhibit significantly higher religious bias scores across all models. This may stem from the strong cultural emphasis on Buddhism in Thailand and Islam in Indonesia, both of which play a central role in their respective societies \citep{pew2017religiouscomposition}. These cultural influences could contribute to a more pronounced religious bias in Thai and Indonesian datasets, making it more challenging to mitigate in multilingual models.

\paragraph{On Gender Bias} \citet{butler2024gender} argues, "\textit{If gender itself is realized through grammatical conventions [...], then at the most fundamental epistemological level, the transformation of gender must involve a challenge to these grammatical structures}." Russian exhibits stronger gender bias compared to other languages, likely due to its extensive gender-based inflection in nouns, adjectives, and verbs, \emph{e.g.}, \textcyr{он говорил} (he spoke) uses the masculine form of the verb, while \textcyr{она говорила}(she spoke) uses the feminine form. This gender marking is not optional—every time a verb is used, it must agree with the subject’s gender. The presence of explicit grammatical gender markers can reinforce gender bias in models by systematically encoding gendered distinctions.
In contrast, Thai and Indonesian, which lack grammatical gender distinctions, do not show a clear pattern of gender bias. In Thai, for instance, (\textit{khao}) can refer to both “he” and “she”, while in Indonesian, \textit{dia} serves as a gender-neutral pronoun, reducing the likelihood of systematic gender bias in models.
This may contribute to lower gender bias compared to Russian, as gender distinctions in Thai are primarily does not affect verbs, adjectives and nouns.
These findings suggest that linguistic structure and cultural norms shape the way bias is encoded in multilingual models.

English, despite having gendered pronouns, lacks the pervasive grammatical gender markers found in Russian, making its gender bias less structurally embedded. These findings suggest that linguistic structures play a crucial role in shaping how gender bias is encoded in multilingual models, with languages like Russian demonstrating a stronger inherent bias due to their grammatical framework, whereas in others, such as Chinese, Thai, and Indonesian, the effects are less obvious.

\paragraph{Key Takeaways}
\begin{itemize}
    
    \item Multilingual models without proper bias mitigation tend to exhibit higher biases in non-English, particularly low-resource languages, emphasizing the need for more diverse and representative training data.
    \item Model performance in a given language does not always correlate with its bias score; in some cases, higher performance is associated with stronger biases, while models with poor language support may produce low bias scores simply due to unintelligible or generic outputs.
    \item Recent LLMs show superior performance on bias in both English and non-English languages, indicating that the large scale might have provided the effectiveness of cross-lingual alignment.
    \item The disparity in bias scores across languages suggests that bias is not solely a result of training data quantity but may also be affected linguistic complexity, script differences, and cultural contexts. This has been observed in prior studies \citep{blodgett-etal-2020-language}, where language-specific biases emerge due to imbalanced representation in training data.

\end{itemize}

\section{Cross-lingual Transfer Debiasing}

We have first examined how biases are presented across languages, and now we proceed to investigate whether other languages can benefit from English corpus-based debiasing methods. We use XLM-RoBERTa in this case as an example because there is a significant gap between English and other languages in terms of both bias severity and representational alignment. This discrepancy makes XLM-RoBERTa an illustrative case for investigating whether debiasing techniques developed for English can transfer effectively to other languages. To quantify the effect of debiasing techniques, we calculated the degree of bias reduction for each debiasing method as the relative percentage reduction:

\begin{equation}
\text{Reduction}=\frac{\mathbb{NBS} - \mathbb{NBS}'}{\mathbb{NBS}}  \times 100\%
\end{equation}

where $\mathbb{NBS}$ is baseline XLM-RoBERTa bias score, and $\mathbb{NBS}'$ is Bias score after applying mitigation.

\subsection{Data}

Following the similar experiments done by \citet{reusens2023investigating}, we select our debiasing experiment data from the Wikipedia dataset \cite{meade2022empirical}, which contains cleaned multilingual full-text Wikipedia articles. 
For the purpose of this research, we selected 10\% of the data from Wikipedia's extensive database for experimentation. Through this approach, we obtained unsupervised data from 514,084 articles as a sufficient sample size for our bias assessment.

\subsection{Methods}
We adpoted three key strategies: \texttt{CDA} \citep{zmigrod2019counterfactual,webster2020measuring}, \texttt{DO} and \texttt{SenDeb} \citep{liang2020towards}. Beyond the three, there are also other approaches such as \texttt{Self-Debias}, \texttt{INLP} and \texttt{DR}. However,  \texttt{Self-Debias} is a post-hoc text generation debiasing procedure and cannot be used as a debiasing technique for downstream natural language understanding tasks; \texttt{INLP}, \texttt{DR} and \texttt{SenDeb} are all projection-based debiasing techniques, therefore we only chose to experiment with \texttt{SenDeb}. 
For a complete methodological-level description, please refer to appendix \ref{DebiasingMethodDetail}.

\begin{itemize}

    \item Counterfactual Data Augmentation (\textbf{\texttt{CDA}}) generates counterfactual samples to balance the dataset. In our experiment, we apply \texttt{CDA} to fine-tune on English data, measure bias shifts in English and four other languages. In the \texttt{CDA} process, for addressing gender bias, we selected common binary replacement pairs, such as \textit{businessman} and \textit{businesswoman}. For racial and religious bias, we used ternary replacement sets, such as \textit{black}, \textit{caucasian}, and \textit{asian}, or \textit{judaism}, \textit{christianity}, and \textit{islam}.
    
    \item Dropout Regularization (\textbf{\texttt{DO}}) randomly \textit{drop} (temporarily remove) certain network nodes during training, forcing the model to learn more generalizable representations. In our experiment, we fine-tune XLM-RoBERTa using \texttt{DO} on English datasets and evaluate across multiple languages. In the \texttt{DO} debiasing process, we adjusted the model's hyperparameters and set the dropout probability of the hidden layers (\texttt{hidden\_dropout\_prob}) to 0.20 and the dropout probability of the attention heads (\texttt{attention\_probs dropout\_prob}) to 0.15.

    \item SentenceDebias (\textbf{\texttt{SenDeb}}) projects bias subspace to biased vectors, and extends debiased word vectors to full sentence representations. In our experiment, we first utilize the dataset processed in \texttt{CDA}, as described above, and then computed the bias subspace for the dataset. 
    For each type of bias, we separately obtained and aligned the corresponding word vector representations. We computed the mean vector for each example, and subtracted it from each word vector to ensure data centering. We applied PCA to the aligned word vector representations and extracted the first principal component as the bias direction. During model inference, we applied projection correction of the bias direction to the last hidden layer to remove bias influence. Specifically, in the model's forward propagation, we perform debiasing operations on the final hidden state of the output layer.

\end{itemize}

\subsection{Results \& Analysis}

\label{sec: debias_reuslt}

Table \ref{tab:BSMitigationResult} shows the percentage reduction in bias for different languages using \texttt{CDA}, \texttt{DO}, and \texttt{SenDeb}. Across all languages, the three debiasing methods demonstrated varying degrees of bias reduction, not just in English. This indicates that these debiasing techniques have cross-lingual applicability and effectiveness.

\begin{table}
    \centering
    \footnotesize
    \begin{tabular}{lccc}
        \hline
        \textbf{Language} & \textbf{\texttt{CDA}}(\%) & \textbf{\texttt{DO}}(\%) & \textbf{\texttt{SenDeb}}(\%) \\ \hline
        English  & -12.69 & -9.55 & -22.42 \\
        Chinese & -5.29 & -2.67 & -37.96 \\ 
        Russian  & -6.30 & -4.96 & -23.86 \\ 
        Thai & -3.86 & -4.61 & -34.33 \\ 
        Indonesian  & -5.06 & -5.13 & -22.93 \\ \hline
    \end{tabular}
    \caption{Debias Method Results across Different Languages with English Data}
    \label{tab:BSMitigationResult}
\end{table}

\texttt{CDA} showed particularly notable results in English. By training the XLM-RoBERTa model with English data, the model
also showed bias reduction effects in other languages, but the overall effect remained primarily focused on English. 
\texttt{DO} showed relatively stable performance across languages, and its effect was most significant in English, while other languages were also positively impacted, demonstrating some debiasing capability. 
\texttt{SenDeb}  showed the highest bias reduction effect across all languages. Surprisingly \texttt{SenDeb} did not show its best performance in English, unlike \texttt{CDA} and \texttt{DO}.
Instead, it performed excellently across all languages, with the highest degrees of bias reduction in Thai and Chinese. As a direct debiasing technique, the \texttt{SenDeb} method might share a common bias subspace across languages. Therefore, this method was able to reduce bias regardless of languages. 
This could potentially be explained by the property of the word vector space, where vector relationships can approximate complex semantic and lexical relationships through linear operations \citep{drozd-etal-2016-word}. In this way, word vectors can not only represent semantic information of words but also achieve analogical reasoning. Although the model was not specifically trained for such tasks, this relational structure of word vectors emerged naturally. In our experiments, we hypothesize that bias-level information, as relational properties, can be largely retained across languages, and overlaps in multilingual contexts. 

This strong result calls back to previous work \citep{chang2022geometrymultilinguallanguagemodel} that investigates how multilingual language models maintain a shared multilingual representation space while still encoding language-sensitive information in each language, by having language-sensitive and language-neutral axes naturally emerged within the representation space. For example, vector differences can represent semantic relationships such as gender and race to certain vector directions, and a bias vector subspace in English, when applied to other languages, can be seen as an approximation of the bias vector subspace of that language's own.
This means that the bias subspace calculated from English can be easily applied to other languages, thereby providing useful tool for cross-lingual bias reduction.

\section{Conclusion and Future Directions}

This study focused on two main aspects: Evaluating bias in a multilingual setting and demonstrating the cross-lingual transferability of debiasing methods, offering new perspectives for ensuring fairness in multilingual models. 

We proposed a new evaluation metric to enable fair comparisons between different models, constructed a multilingual bias evaluation dataset, consisting of languages with different resource status, and benchmarked major multilingual language models. We have found that for LLMs transfer learning of fairness is much better for non-English language compared to previous models, and suggested that there exists a nuance of social bias when dealing with different linguistic and cultural backgrounds. Our study demonstrates that high-resource language debias methods can be effectively transferred to low-resource languages for bias mitigation. \textbf{\texttt{SenDeb}} emerged as the most effective technique, suggesting that bias subspaces may share cross-linguistic properties, enabling cross-language debiasing. This finding opens new possibilities for developing universal debiasing methods across diverse languages.

Based on this work, more could be explored to further advance fairness in multilingual NLP, to improve inclusivity and ethical integrity of NLP worldwide:

\begin{itemize}
    \item Expand and diversify Bias Evaluation Datasets, covering more languages and cultural contexts to enhance the comprehensiveness of bias evaluation. 
    \item Conduct more detailed studies on different bias types and develop more language-aware debias methods. Also explore debiasing techniques more specifically tailored to causal LMs.
    \item Explain why bias subspace are shared mutually in various languages on a interpretability level and design better approach to align them. Develop a universal debiasing method that works equally well on a pan-linguistic scale.

\end{itemize}

\section*{Limitations}

Our bias evaluation dataset was constructed through translation, using English—specifically American English—as the cultural and linguistic base. While this enabled efficient multilingual expansion, it risks importing U.S.-centric social norms into other languages, potentially skewing bias evaluations and leading to unfair cross-lingual comparisons. Also, in this work we rely on Google Translate to construct new corpus into due to budget and availability constraints, which restricted our ability to ensure culturally nuanced translations. To address the issue mentioned above, future work should move beyond translation-based methods and invest in building culturally grounded datasets developed natively in each language. Collaborating with native speakers and cultural experts will be key to capturing locally relevant social dynamics and ensuring more accurate, equitable bias evaluations.

The bias evaluation and debias methods in this study were designed for masked language models. While we approximate their application to causal language models (autoregressive transformers), there may be inherent differences in how bias manifests and propagates in these models. 

While our study suggests that bias subspaces share cross-linguistic properties, the extent to which this holds across all languages remains an open question. The effectiveness of debiasing methods may vary for languages with fundamentally different linguistic structures or those underrepresented in pretraining corpora. Further research is needed to validate and refine our findings across a wider linguistic and typological spectrum.




\bibliography{bib/ref.bib}

\appendix

\section{$\mathbb{NBS}$: A Definition}

\label{sec:NBS_definition}

This section  provides a theoretical explanation of how social bias in language models can be evaluated. \citet{nadeem2020stereoset} first proposed that bias in language models can be assessed using Masked Language Models (MLMs) \cite{devlin2019bert}, where bias is measured by predicting the probability of masked words. The specific measurement method is as follows:

Given a sentence $s$, $s$ contains a specific social attribute (\emph{e.g.}, \textit{\textbf{Mr.} Li is a university professor.}), we can modify the words associated with that attribute (\emph{e.g.}, \textit{\textbf{Mrs.} Li is a university professor.}). Let 
$M = \{m_{0},\ldots,m_{n}\}$ represent the modified words (\textit{\textbf{Mr.}, \textbf{Mrs.}}),
$U = \{ u_0,\ldots,u_l\}$ represent the unchanged words (\textit{\textbf{Li}, \textbf{is}, \textbf{a}...}), then we have the modified sentence
$s_i = U \cup m_i$.

Assuming the masked language model has parameters 
$\theta$, we can measure the model’s bias towards specific social attributes by masking the words in 
$M$ and predicting their probabilities. By comparing the probabilities for different words 
$m_i\in M$, we can reasonably assess the probability of $s_i$ in the language model:

\begin{equation}
   P(s_i) = P(m_i \space | \space  U; \theta) 
\end{equation}

However, \citet{nangia2020crows} pointed out that the probability $P(m_i)$ would also affect model's prediction. This frequency bias does not necessarily indicate social bias in the language model itself. To address this issue, they proposed probability score $\mathbb{PS}$ evaluating the probability of unchanged words given the modified words by applying pseudo-log-likelihood estimation \citep{wang2019bert}. For modified sentence $s_i$, words in $U$ are masked one at a time until all 
$u_j$ have been masked:

\begin{equation}
   \mathbb{PS}({s_i}) = \sum_{j=0}^{|U|} \log P(u_j \in U | U_{\backslash u_j}; m_i; \theta) 
\end{equation}


This score approximates the true conditional probability, measuring how strongly a model assigns higher likelihoods to stereotypical sentences.

While \citet{nangia2020crows}'s method allows for intra-model comparisons, it does not facilitate bias comparisons across different models. Language models may have different architectures and training datasets, leading to varying internal weight distributions that affect their predictions under the same evaluation conditions.

To address this, this paper proposes normalizing each model's predictions by computing the average bias prediction score across different social attributes, using the formula below (\ref{equ:13}). This method enables fair and consistent bias comparison across models and provides a comprehensive bias evaluation framework.

Because in the scope of this research we only look at modification where the modification is binary, we would simply the notation so as the original sentence is $s$ and the modified is $\bar{s}$, 

\begin{equation}
\label{equ:13}
 W_{avg} = \frac{1}{n} \sum_{l\in lang} \cdot \frac{1}{N}\sum_{k=1}^{N} \frac{\left| \mathbb{PS}({s_{l,k}}) + \mathbb{PS}({\bar{s}_{l,k}}) \right|}{2}
\end{equation}

\begin{equation}
\label{equ:only}
\begin{split}
\mathbb{NBS}(\theta) & = \frac{1}{W_{avg}}\cdot \frac{1}{N} \sum_{k=1}^{N} \left| \mathbb{PS}({s_{l,k}}) - \mathbb{PS}({\bar{s}_{l,k}}) \right| \\ \\
           & = 2n \cdot 
           \frac
           {\sum_{k=1}^{N} \left| \mathbb{PS}({s_{l,k}}) - \mathbb{PS}({\bar{s}_{l,k}}) \right|}
           {\sum_{l\in lang} \sum_{k=1}^{N} \left| \mathbb{PS}({s_{l,k}}) + \mathbb{PS}({\bar{s}_{l,k}}) \right|}
\end{split}
\end{equation}
where $n = \left| lang \right| $.

In the analysis, bias evaluations are conducted using this $\mathbb{NBS}(\theta)$ metric. The closer $\mathbb{NBS}$ is to 0, the lower the bias in the model. If $\mathbb{NBS}$ = 0, it indicates the model treats the two terms equally and exhibits no intrinsic social bias.

\section{Model Details}
\label{sec:models}

In this study we selected three pre-trained models: mBERT, XLM-RoBERTa, BLOOM, XGLM, Qwen and LLaMA. These models are widely used particularly in multilingual tasks.

\textbf{mBERT} \citep{devlin2019bert} is a multilingual version of BERT trained on Wikipedia data in 104 languages.
Similar to English BERT, mBERT utilizes Masked Language Modeling (MLM) and Next Sentence Prediction (NSP) tasks for training.
mBERT does not require language-specific adaptation, making it directly applicable to text analysis in multiple languages.
In this experiment, we used the \texttt{google-bert/bert-base-multilingual-cased} version to evaluate bias across different languages.

\textbf{XLM-RoBERTa} \cite{conneau2020unsupervised} is a cross-lingual language model based on the RoBERTa architecture, trained on 100 languages.
Unlike mBERT, XLM-RoBERTa uses a larger training dataset and longer training cycles, while removing the NSP task and relying solely on MLM.
This allows XLM-RoBERTa to perform better in cross-lingual understanding tasks.
Our experiments utilized the \texttt{FacebookAI/xlm-roberta-base} model.


\textbf{XGLM} \citep{lin2022fewshotlearningmultilinguallanguage} is a multilingual autoregressive language model designed to facilitate few-shot learning across multiple languages. Trained on a balanced corpus spanning 30 diverse languages and totaling 500 billion sub-tokens, XGLM aims to provide robust cross-linguistic generalization without requiring extensive task-specific finetuning. It follows a decoder-only Transformer architecture, making it well-suited for text generation and language modeling tasks. XGLM has demonstrated strong few-shot performance, highlighting its ability to adapt to new tasks with minimal supervision. In this experiment, we used the \texttt{facebook/xglm-564m} version to evaluate bias across different languages.

\textbf{Gemma 3} \citep{gemmateam2024gemmaopenmodelsbased} models are available in various parameter sizes, including 1B, 4B, 12B, and 27B, and are designed for multimodal text and image processing. Gemma 3 follows an autoregressive Transformer architecture and supports a large 128K context window, making it suitable for tasks such as question answering, summarization, and reasoning. The model incorporates supervised fine-tuning (SFT) and reinforcement learning with human feedback (RLHF) to enhance alignment with human preferences and safety considerations. With multilingual support spanning over 140 languages, Gemma 3 is optimized for global usability and can be further fine-tuned for domain-specific applications. In this experiment, we used the \texttt{google/gemma-3-1b-pt} version to evaluate bias across different languages.

\textbf{Qwen2.5} \citep{qwen2025qwen25technicalreport} is the latest iteration in the Qwen series of large language models, offering improvements in knowledge retention, coding, and mathematical reasoning. Qwen2.5 models range from 0.5B to 72B parameters and provide enhanced instruction-following, long-text generation, and structured data comprehension. The models support over 29 languages, including Chinese, English, Russian and Indonesian. However, there is no report of Qwen 2.5 supporting Thai language. Qwen2.5 employs a Transformer architecture with RoPE, SwiGLU, and RMSNorm, along with Grouped-Query Attention (GQA) for efficiency. In this experiment, we used the \texttt{Qwen/Qwen2.5-0.5B} version to evaluate bias across different languages.

\textbf{LLaMA 3} \citep{grattafiori2024llama3herdmodels} models are available in 1B and 3B parameter sizes and are optimized for multilingual dialogue, agentic retrieval, and summarization tasks. LLaMA 3 follows an autoregressive Transformer architecture and benefits from supervised fine-tuning (SFT) and reinforcement learning with human feedback (RLHF) to align with human preferences for helpfulness and safety. It supports a range of languages, with potential for further fine-tuning on additional languages. In this experiment, we used the \texttt{meta-llama/Llama-3.2-1B} version to evaluate bias across different languages.

\section{Debiasing Method Details}

\label{DebiasingMethodDetail}

\subsection{\texttt{CDA}}
Counterfactual Data Augmentation (\textbf{\texttt{CDA}}) \citep{zhao-etal-2018-gender,zmigrod2019counterfactual,webster2020measuring} is a bias mitigation technique that generates counterfactual samples to balance the dataset. It involves supplementing training data with modified sentences and evaluating the impact on bias reduction in low-resource language datasets. The process consists of the following steps:
\begin{enumerate}
    \item Duplicating sentences that contain predefined biased attribute words.
    \item Swapping biased attributes with their counterfactual counterparts (\emph{e.g.}, replacing \textit{he} with \textit{she}).
    \item Fine-tuning the model with the augmented dataset to reduce bias
    
\end{enumerate}

Previous studies \cite{webster2020measuring} have shown that training English models (\emph{e.g.}, ALBERT and BERT) on \texttt{CDA}-augmented datasets can significantly reduce bias. However, some research \cite{reusens2023investigating} found that fine-tuning mBERT on English datasets actually increased bias in French and German. In our experiment, we apply \texttt{CDA} fine-tuning on English data, measure bias changes in English and four other languages, and determine whether English-based bias mitigation can be effectively transferred via multilingual learning.

\subsection{\texttt{DO}}
Dropout Regularization (\textbf{\texttt{DO}}) is a commonly used deep learning technique to prevent overfitting. The idea is to randomly \textit{drop} (temporarily remove) certain network nodes during training, forcing the model to learn more generalizable representations. Since \texttt{DO} disrupts word association patterns in attention mechanisms, previous research \citep{webster2020measuring} hypothesized that \texttt{DO} could also reduce gender and other types of bias. Studies found that \texttt{DO} fine-tuning effectively reduced bias in ALBERT and BERT, without modifying training data distribution or making explicit assumptions about bias patterns. Additionally, \cite{reusens2023investigating} reported that \texttt{DO} fine-tuning in English reduced bias by 10\% on French models.

In our experiment, we fine-tune XLM-RoBERTa using \texttt{DO} on English datasets and evaluate its impact on bias reduction across multiple languages.

\subsection{\texttt{SenDeb}}
\begin{algorithm*}
\caption{SenDeb: Sentence Representation Vector Debiasing Algorithm}
\begin{algorithmic}
\State Initialize the sentence (pretrained) encoder $M_\theta$.
\State Define bias types (e.g., binary gender: male $g_m$ and female $g_f$).
\State Design the bias attribute lexicon $D = \{ (w_1^{(i)}, \ldots, w_d^{(i)}) \}^m_{i=1}$.
\State $S = \bigcup_{i=1}^m \text{C\small{ONTEXTUALIZE}}(w_1^{(i)}, \ldots, w_d^{(i)}) = \{ (s_1^{(i)}, \ldots, s_d^{(i)}) \}^n_{i=1}$ \Comment{Integrate words into sentences}
\For{$j \in [d]$}
    \State $R_j = \{ M_\theta(s_j^{(i)}) \}^n_{i=1}$ \Comment{Obtain sentence vectors}
\EndFor
\State $V = \text{PCA}_k \left( \bigcup_{j=1}^d \bigcup_{w \in R_j} ( w - \mu_i ) \right)$ \Comment{Compute the bias subspace}
\For{each new sentence vector $h$}
    \State $h_V = \sum_{j=1}^k \langle h, v_j \rangle v_j$ \Comment{Compute projection onto the bias subspace}
    \State $\hat{h} = h - h_V$ \Comment{Subtract the projection}
\EndFor
\end{algorithmic}
\end{algorithm*}
SentenceDebias (\textbf{\texttt{SenDeb}})
\citep{liang2020towards} is a projection-based debiasing technique that extends debiased word vectors to full sentence representations. Previous research on bias mitigation tends to operate at the word level. However, supervised datasets are limited by vocabulary size \citep{bolukbasi2016man}, whereas the number of possible sentences is infinite, making it extremely difficult to precisely characterize bias-free sentences. Therefore, our approach converts these words into sentences to obtain feature representations from a pretrained sentence encoder. The following subsections describe the method used to address this problem. The specific implementation steps are as follows:
\begin{enumerate}
    \item \textbf{Defining Bias Attributes}: For example, when characterizing gender bias, we use word pairs such as \textit{(male, female)} to represent gender. Each tuple should consist of words that are semantically equivalent except for the bias attribute. Typically, for $d$-class bias attributes, the word pairs form a dataset $D = \{(w_1^{(i)},\ldots, w_d^{(i)}))\}^m_{i=1}$ with $m$ entries, where each entry $(w_1, \ldots, w_d)$ is a $d$-tuple.
    \item \textbf{Computing the Subspace}: There exists a common bias subspace in all possible sentence representations. To accurately estimate this bias subspace, we should use as diverse sentence templates as possible to account for the word’s position in surrounding contexts. In experiments, we retrieve attribute words from a corpus and place them into biased attribute sentences using a \texttt{CDA}-based approach, further obtaining their sentence representations. This results in a significantly expanded biased attribute sentence dataset $S$:
    \begin{equation}
    \begin{split}
        S & = \bigcup_{i=1}^m \text{C\small{ONTEXTUALIZE}}(w_1^{(i)}, \ldots, w_d^{(i)}) \\  
        & = \{ (s_1^{(i)}, \ldots, s_d^{(i)}) \}^n_{i=1} 
    \end{split}
    \end{equation}

    In an encoder $M_\theta$ parameterized by $\theta$, the sentence representation vectors $R_j, j \in [d]$ satisfy $R_j = \{ M_\theta(s_j^{(i)}) \}^n_{i=1}$. Among all sentence representation vectors, we can estimate the bias subspace using Principal Component Analysis (PCA) \citep{abdi2010principal}. Defining $\mu_j = \frac{1}{\left|R_j\right|} \sum_{w \in R_j} w$, and assuming that the first $K$ dimensions of PCA define the bias subspace, the subspace $V = \{v_1,\ldots,v_k\}$ satisfies:
    \begin{equation}
        V = \text{PCA}_k \left( \bigcup_{j=1}^d \bigcup_{w \in R_j} ( w - \mu_i ) \right)
    \end{equation}
    \item \textbf{Removing Subspace Projection}: By removing the projection onto this bias subspace, we can eliminate bias in general sentence representations. Given a sentence representation vector $h$, we first compute its projection $h_V$ onto the bias subspace and then subtract this projection to obtain a vector $\hat{h}$ that is approximately bias-free and orthogonal to the bias subspace.
    \begin{equation}
    h_V = \sum_{j=1}^k \langle h, v_j \rangle v_j
    \end{equation}
    \begin{equation}
    \hat{h} = h - h_V
    \end{equation}
\end{enumerate}

\begin{figure*}[t]
  \centering
  \includegraphics[width=0.55\linewidth]{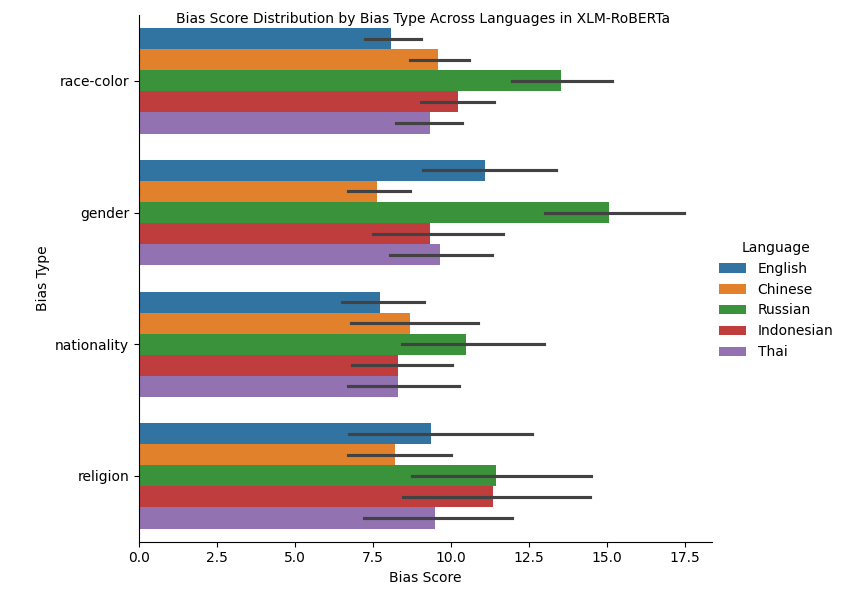}
  \includegraphics[width=0.38\linewidth]{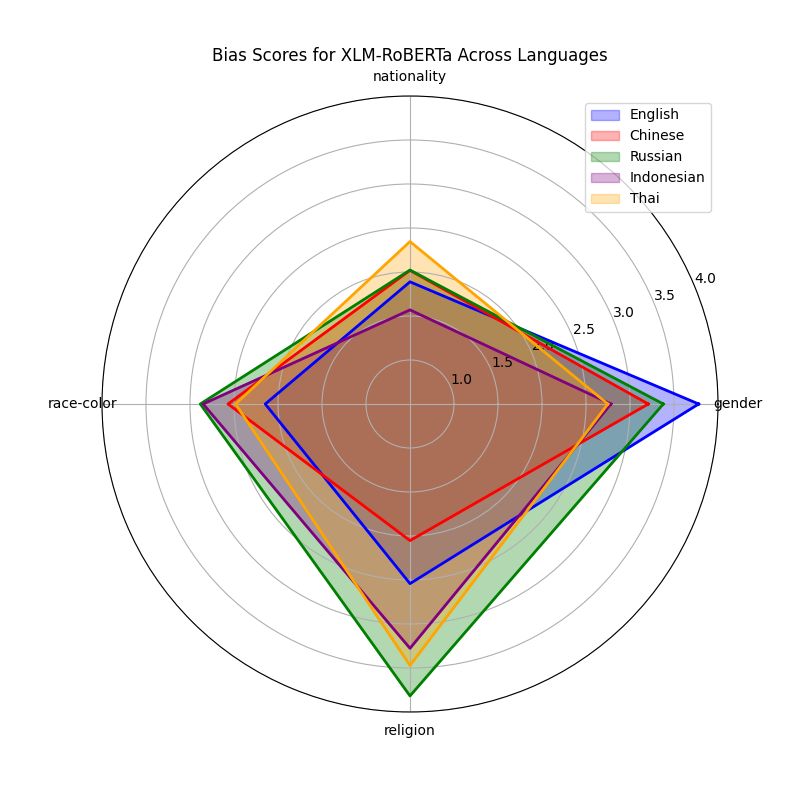}
  \caption{Bias Score Distribution by Bias Type Across Langauges in XLM-RoBERTa as bar charts and radar charts.}
  \label{fig:barandradar}
\end{figure*}


Beyond these three methods, there are also other approaches such as \texttt{Self-Debias}, Iterative Nullspace Projection (\texttt{INLP}), and DensRay (\texttt{DR}). However, since \texttt{Self-Debias} is a post-hoc text generation debiasing procedure, it cannot be used as a debiasing technique for downstream natural language understanding tasks. Furthermore, \texttt{INLP}, \texttt{DR}, and \texttt{SenDeb} are all projection-based debiasing techniques, therefore we only chose to experiment with \texttt{SenDeb}. Through these three methods, we attempted to conduct experiments using multilingual models and report the bias indices before the debiasing experiments and the optimization improvements after the experiments were completed, in order to measure the effectiveness of the debiasing techniques.

\section{Bias Score Visualizations}
\label{sec:appendix-graph}

In this section, we present additional visualizations of the bias scores computed across different models and languages. The figures provide a detailed breakdown of bias distributions for gender, nationality, race-color, and religion. Each plot illustrates how bias manifests in different linguistic contexts, allowing for a comparative analysis of bias trends across multilingual models.

\begin{figure*}[t]
  \centering

    \includegraphics[width=1.03\linewidth]{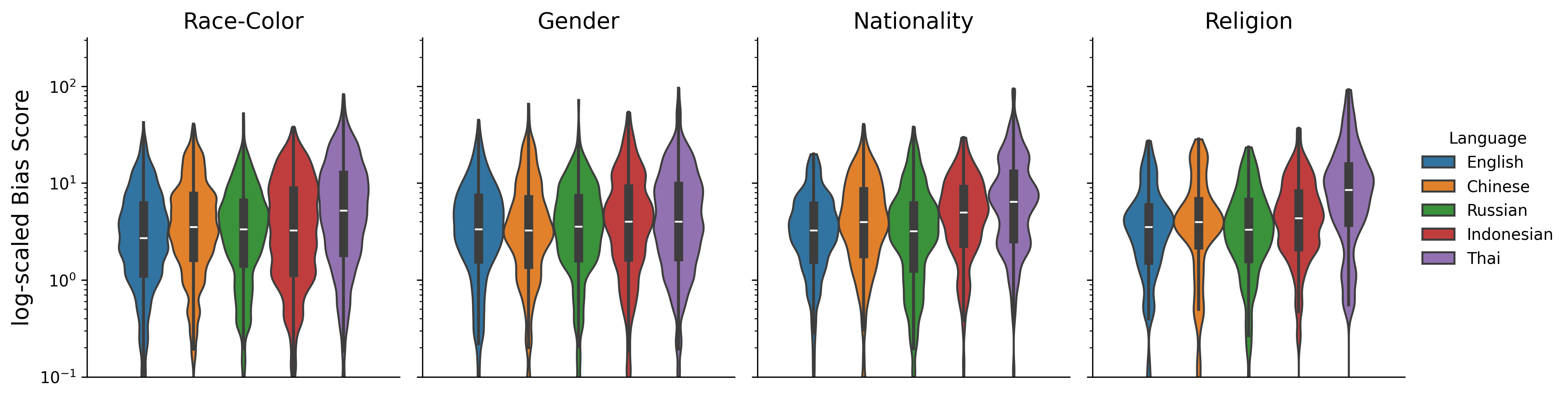} \\[0.7cm]
  \includegraphics[width=1.03\linewidth]{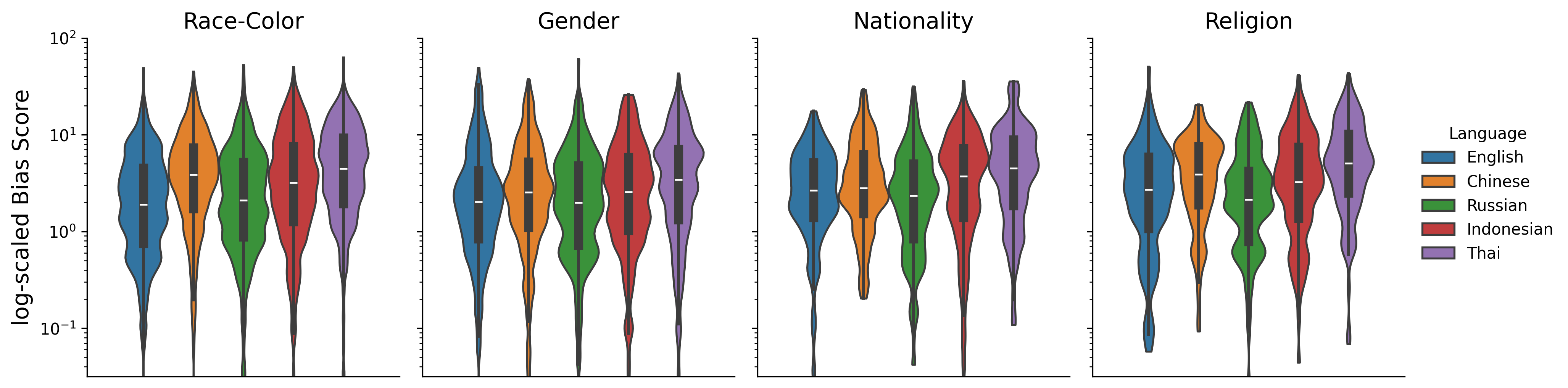} \\[0.7cm]
  \includegraphics[width=1.03\linewidth]{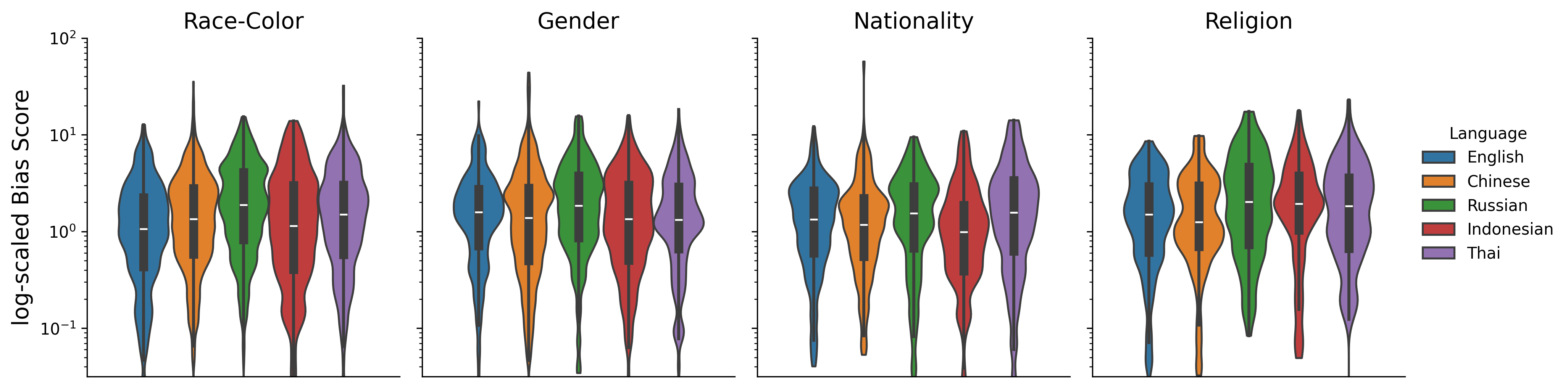} \\[0.7cm]

  \includegraphics[width=1.03\linewidth]{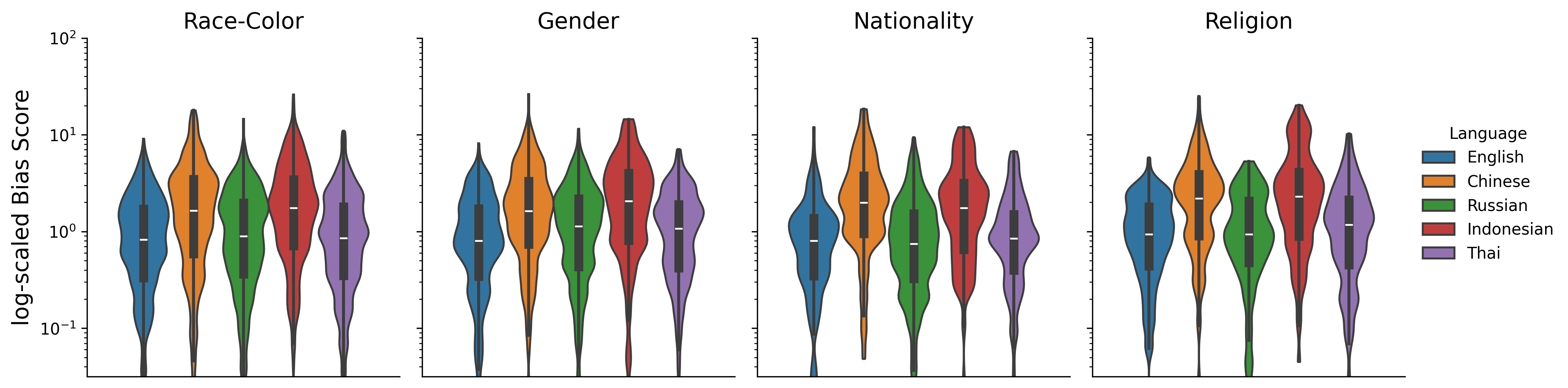} \\[0.7cm]
  \includegraphics[width=1.03\linewidth]{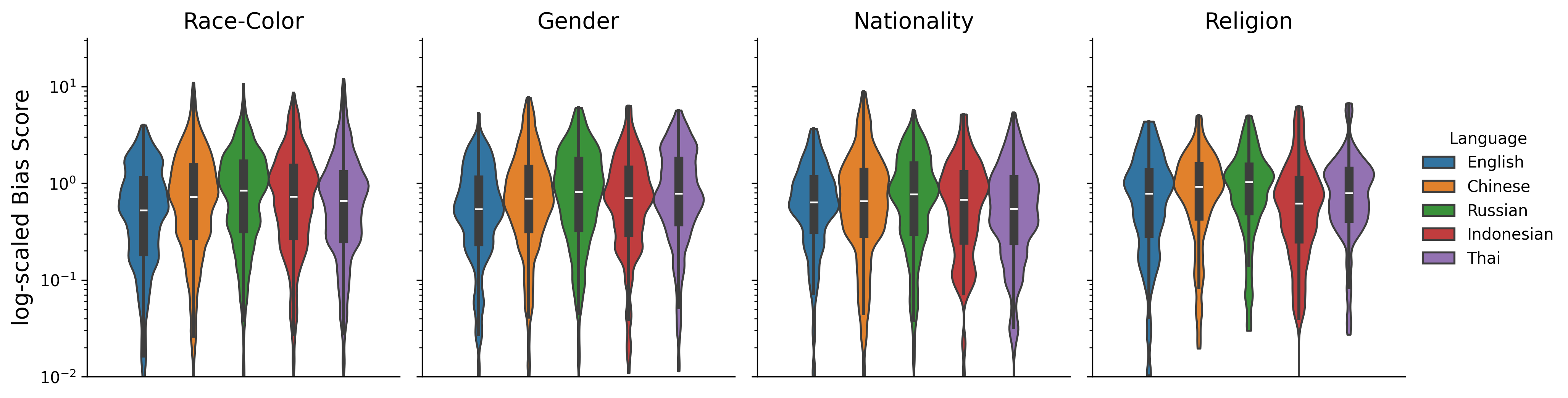} 

  \caption{Bias Score Distribution Across Different Languages and Bias Categories in  \textbf{mBERT}, \textbf{XLM-RoBERTa}, \textbf{XGLM}, \textbf{Qwen 2.5} , \textbf{LLaMA 3}, from top to bottom.}
  \label{fig:violin-xlm}
\end{figure*}

Figures~\ref{fig:barandradar}-\ref{fig:violin-xlm} depict the bias scores for various models: mBERT, XLM-RoBERTa, Qwen 2.5, XGLM and LLaMA 3 The x-axis represents the log-scaled bias score, while the y-axis categorizes bias types across different languages. To enhance readability, we have positioned the legends outside the main plots and adjusted the figure sizes accordingly.

\end{document}